\documentclass{article} 

\usepackage{booktabs}

\usepackage{amssymb,amsmath,amsthm}
\usepackage{bm}

\usepackage{graphicx}
\usepackage{hyperref}

\usepackage[utf8]{inputenc}
\usepackage[T1]{fontenc}
\usepackage{csquotes}

\newcommand{\R}{\mathbb{R}}
\newcommand{\N}{\mathbb{N}}
\newcommand{\effic}{\mathcal{O}}
\DeclareMathOperator*{\argmin}{argmin}
\newcommand{\rmse}{E}

\newcommand{\graph}{\mathcal{G}}
\newcommand{\nodes}{V}
\newcommand{\node}{v}
\newcommand{\lnode}{u}
\newcommand{\rnode}{\node}
\newcommand{\neigh}{\mathcal{N}}
\newcommand{\edges}{E}
\newcommand{\edge}{e}

\newcommand{\lifetime}{\mathcal{T}}
\newcommand{\nodePres}{\psi}
\newcommand{\edgePres}{\rho}

\newcommand{\mat}[1]{\bm{#1}}
\newcommand{\transp}[1]{{ #1 }^T}
\newcommand{\inv}[1]{\frac{1}{#1}}
\newcommand{\eye}{\mat{I}}

\newcommand{\dims}{R}

\newcommand{\timeidx}{t}
\newcommand{\timelim}{T}

\newcommand{\trajlim}{M}
\newcommand{\trajidx}{j}

\newcommand{\inspace}{\mathcal{X}}
\newcommand{\point}{x}
\newcommand{\inmat}{\mat{X}}

\newcommand{\outpoint}{y}
\newcommand{\outmat}{\mat{Y}}
\newcommand{\dataidx}{i}
\newcommand{\datalim}{N}

\newcommand{\coeff}{\alpha}
\newcommand{\fun}{f}

\newcommand{\kernel}{k}
\newcommand{\kernelvec}{\vec{k}}
\newcommand{\kernelmat}{\mat{K}}
\newcommand{\kernelfeat}{\phi}

\newcommand{\kernelspace}{\mathcal{Y}}
\newcommand{\rbfband}{\psi}

\newcommand{\simi}{s}
\newcommand{\simimat}{\mat{S}}

\newcommand{\diss}{d}
\newcommand{\avgDiss}{\bar d}

\newcommand{\dissmat}{\mathrm{D}}
\newcommand{\dissfeat}{\phi}
\newcommand{\dissspace}{\mathcal{Y}}

\newcommand{\prob}{P}
\newcommand{\dens}{p}
\newcommand{\nDens}{\mathcal{N}}
\newcommand{\nMean}{\mu}
\newcommand{\nDev}{\sigma}
\newcommand{\nDevNoise}{\tilde \nDev}
\newcommand{\nCov}{\mat{C}}

\newcommand{\Prior}{\mat{\Theta}}
\newcommand{\prior}{\theta}
\newcommand{\gpcoeff}{\gamma}
\newcommand{\clustlim}{C}
\newcommand{\clustidx}{c}
\newcommand{\clustweight}{\beta}
\newcommand{\rbcm}{\mathrm{rBCM}}

\theoremstyle{definition}
\newtheorem{definition}{Definition}
\theoremstyle{plain}
\newtheorem{theorem}{Theorem}

\begin{document}

\title{Time Series Prediction for Graphs in Kernel and Dissimilarity Spaces%
\thanks{Funding by the DFG under grant number HA 2719/6-2 and the CITEC center of 
excellence (EXC 277) is gratefully acknowledged.}
\thanks{This contribution is an extension of the work presented at ESANN 2016 under the title
\enquote{Gaussian process prediction for time series of structured data} \cite{Paassen2016ESANN}.}}

\author{Benjamin Paaßen \and Christina Göpfert \and Barbara Hammer}

\date{This is a preprint of the publication \cite{Paassen2017NPL} as provided by the authors.
The original can be found at the DOI \href{https://doi.org/10.1007/s11063-017-9684-5}{10.1007/s11063-017-9684-5}.}

\maketitle

\pagestyle{myheadings}
\markright{Preprint of the publication \cite{Paassen2017NPL} as provided by the authors.}

\begin{abstract}
Graph models are relevant in many fields, such as distributed computing, intelligent tutoring systems or
social network analysis. In many cases, such models need to take changes in the graph structure into
account, i.e.\ a varying number of nodes or edges. Predicting such changes within graphs can
be expected to yield important insight with respect to the
underlying dynamics, e.g.\ with respect to user behaviour. However, predictive techniques in the
past have almost exclusively focused on single edges or nodes. In this contribution, we attempt to predict
the future state of a graph as a whole.
We propose to phrase time series prediction as a regression problem and apply dissimilarity- or kernel-based
regression techniques, such as 1-nearest neighbor, kernel regression and Gaussian process regression,
which can be applied to graphs via graph kernels.
The output of the regression is a point embedded in a pseudo-Euclidean space, which can be analyzed
using subsequent dissimilarity- or kernel-based processing methods. We
discuss strategies to speed up Gaussian Processes regression from cubic to linear time and
evaluate our approach on two well-established theoretical models of graph evolution as well as
two real data sets from the domain of intelligent tutoring systems. We find that simple regression methods,
such as kernel regression, are sufficient to capture the dynamics in the theoretical models,
but that Gaussian process regression significantly improves the prediction error for real-world data.
\end{abstract}

\section{Introduction}

To model connections between entities, graphs are oftentimes the method of choice,
e.g.\ to model traffic connections between cities \cite{Papageorgiou1990}, data lines between
computing nodes \cite{Casteigts2012}, communication between people in social networks
\cite{Nowell2007}, or the structure of a student's solution to a learning task in an intelligent tutoring
system \cite{Mokbel2013,Paassen2016}. In all these examples, nodes as well as connections change significantly
over time. For example, in traffic graphs, the traffic load changes significantly over the course
of a day, making optimal routing a time-dependent problem \cite{Papageorgiou1990};
in distributed computing, the distribution of computing load and
communication between machines crucially depends on the availability and speed of connections
and the current load of the machines, which changes over time \cite{Casteigts2012};
in social networks or communication networks new users may enter the network,
old users may leave and the interactions between users may change rapidly \cite{Nowell2007};
in intelligent tutoring systems, students change their solution over time to get closer to
a correct solution \cite{Koedinger2013,Mokbel2013}. In all these
cases it would be beneficial to predict the next state of the graph in question, because it provides
the opportunity to intervene before negative outcomes occur, for example by re-routing traffic,
providing additional bandwidth where required or to provide helpful hints to students.

Traditionally, predicting the future development based on knowledge of the past is the topic of
\emph{time series prediction}, which has wide-ranging applications in physics, sociology, medicine,
engineering, finance and other fields \cite{Sapankevych2009,Shumway2013}. However, classic models
in time series prediction, such as ARIMA, NARX, Kalman filters, recurrent networks
or reservoir models focus on vectorial data representations, and they are not equipped to handle 
time series of graphs \cite{Shumway2013}. Accordingly, past work on predicting changes in graphs
has focused on simpler sub-problems which can be phrased as classic problems, e.g.\
predicting the overall load in an energy network \cite{Ahmad2014} or predicting the appearance of
single edges in a social network \cite{Nowell2007}.

In this contribution, we develop an approach to address the time series prediction problem for
graphs, which we frame as a regression problem with structured data as input \emph{and as output}.
Our approach has two key steps: First, we represent graphs via pairwise kernel values, which are
well-researched in the scientific literature
\cite{Aiolli2015,Borgwardt2005,da_san_martino_ordered_2016,DaSanMartino2010,Feragen2013,shervashidze_weisfeiler-lehman_2011}.
This representation embeds the discrete set of graphs in a smooth kernel space.
Second, within this space, we can apply similarity- and kernel-based regression methods,
such as nearest neighbor regression, kernel regression \cite{Nadaraya1964} or Gaussian processes \cite{Rasmussen2005}
to predict the next position in the kernel space given the current position.
Note that this does \emph{not} provide us with the graph that corresponds to the predicted point in
the kernel space. Indeed, identifying the corresponding graph in the primal space is a
\emph{kernel pre-image problem} which is in general hard to solve \cite{bakir_learning_2003,bakir_learning_2004,Kwok2004}.
However, we will show that this data point can still be analyzed with subsequent kernel- or dissimilarity based methods.

If the underlying dynamics in the kernel space can not be captured by a simple regression scheme,
such as 1-nearest neighbor or kernel regression, more complex models, such as Gaussian processes (GPs),
are required. However, GPs suffer from a relatively high computational complexity due to a
kernel matrix inversion. Fortunately, Deisenroth and Ng have suggested a simple strategy to permit predictions in linear time,
namely distributing the prediction to multiple Gaussian processes, each of which handles only a
constant-sized subset of the data \cite{Deisenroth2015}.
Further, pre- and post-processing of kernel data - e.g.\ for eigenvalue correction -
usually requires quadratic or cubic time. This added complexity can be avoided
using the well-known Nyström approximation as investigated by \cite{Gisbrecht2015}.

The key contributions of our work are: First, we provide an integrative overview of seemingly disparate
threads of research which are related to time-varying graphs. Second, we provide a scheme for time
series prediction in dissimilarity and kernel spaces. This scheme is compatible with explicit vectorial embeddings,
as are provided by some graph kernels \cite{Borgwardt2005,da_san_martino_ordered_2016},
but does not require such a representation.
Third, we discuss how the predictive result, which is a point in an implicit kernel feature space, can be analyzed using
subsequent kernel- or dissimilarity-based methods.
Fourth, we provide an efficient realization of our prediction pipeline for Gaussian processes in linear time.
Finally, we evaluate our proposed approaches on two theoretical and two practical data sets.

We start our investigation by covering related work and introducing notation
to describe dynamics on graph data. Based on this work, we develop our proposed
approach of time series prediction in kernel and dissimilarity spaces. We also
discuss speedup techniques to provide predictions in linear time. Finally, we
evaluate our approach empirically on four data sets: two well-established theoretical models of
graph dynamics, namely the Barabasi-Albert model \cite{Barabasi1999} and Conway's \emph{Game of Life} \cite{Gardner1970},
as well as two real-world data sets of Java programs, where we try to predict the next step
in program development \cite{Mokbel2013,Mokbel2014}. We find that for the theoretical models,
even simple regression schemes may be sufficient to capture the underlying dynamics,
but for the Java data sets, applying Gaussian processes significantly reduces the prediction error.

\section{Background and Related Work}
\label{sec:background}

Dynamically changing graphs are relevant in many different fields, such as traffic \cite{Papageorgiou1990},
distributed computing \cite{Casteigts2012}, social networks \cite{Nowell2007} or intelligent tutoring systems \cite{Koedinger2013,Mokbel2013}.
Due to the breadth of the field, we focus here on relatively general concepts which can be
applied to a wide variety of domains. We begin with two formalisms to model dynamics
in graphs, namely time-varying graphs \cite{Casteigts2012}, and sequential dynamical systems \cite{Barrett2000}.
We then turn towards our research question: \emph{predicting} the dynamics in graphs. This has 
mainly been addressed in the domain of social networks under the umbrella of link prediction \cite{Nowell2007,Yang2015},
as well as in models of graph growth \cite{Goldenberg2010}. Finally, as preparation for
our own approach, we discuss graph kernels and dissimilarities as well as prior work on kernel-based
approaches for time series prediction.

\subsection{Models of Graph Dynamics}

\paragraph{Time-Varying Graphs:} Time-varying graphs have been introduced by Casteigts and colleagues in an effort to integrate different
notations found in the fields of delay-tolerant networks, opportunistic-mobility networks or social
networks \cite{Casteigts2012}. The authors note that, in all these domains, graph topology changes over time and the
influence of such changes is too severe as to model them only in terms of system anomalies. Rather,
dynamics have to be regarded as an \enquote{integral part of the nature of the system} \cite{Casteigts2012}.
We revisit a slightly varied version of the notation developed in their work.

\begin{definition}[Time-Varying Graph \cite{Casteigts2012}]
A \emph{time-varying graph} is defined as a five-tuple $\graph = (\nodes, \edges, \lifetime, \nodePres, \edgePres)$
where
\begin{itemize}
	\item $\nodes$ is an arbitrary set called \emph{nodes},
	\item $\edges \subseteq \nodes \times \nodes$ is a set of node tuples called \emph{edges},
	\item $\lifetime = \{\timeidx : \timeidx_0 \leq \timeidx \leq \timelim, \timeidx \in \N\}$ is called \emph{lifetime} of the graph,
	\item $\nodePres : \nodes \times \lifetime \to \{0, 1\}$ is called \emph{node presence function},
	and node $\node$ is called \emph{present} at time $\timeidx$ if and only if $\nodePres(\node, \timeidx) = 1$, and
	\item $\edgePres : \edges \times \lifetime \to \{0, 1\}$ is called \emph{edge presence function},
	and edge $\edge$ is called \emph{present} at time $\timeidx$ if and only if $\edgePres(\edge, \timeidx) = 1$.
\end{itemize}
\end{definition}

\begin{figure}
\includegraphics[width=\textwidth]{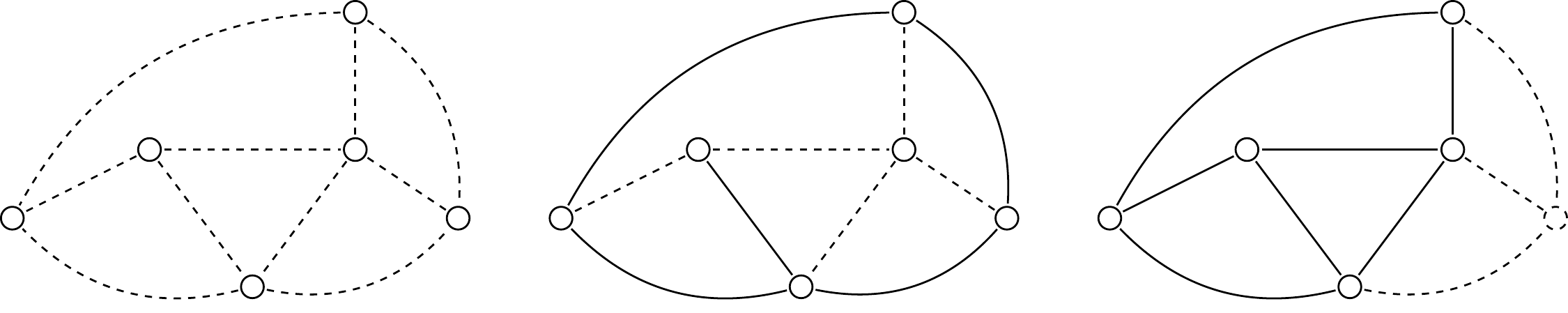}
\caption{An example of a time-varying graph modeling a public transportation graph drawn for
three points in time: night time (left), the early morning (middle) and mid-day (right).}
\label{fig:vbb_example}
\end{figure}

In figure~\ref{fig:vbb_example}, we show an example of a time-varying graph modeling the connectivity
in simple public transportation graph over the course of a day.
In this example, nodes model stations and edges model train connections between stations.
In the night (left), all nodes may be present but no edges, because no lines are active yet. During the early morning
(middle), some lines become active while others remain inactive. Finally, in mid-day (right), all lines
are scheduled to be active, but due to a disturbance - e.g.\ construction work - a station is closed
and all adjacent connections become unavailable. Note that in this example, all nodes, or stations,
are assumed to be known in advance. If not all nodes are known in advance, e.g.\ a long-term public transportation model
where new stations may be built in the future, the definition of time-varying graphs can be extended to allow an infinite
node set $\nodes$ with the restriction that only a finite number of nodes may be present at each time.
This is also relevant for modelling social networks or student solutions in intelligent tutoring systems,
where the introduction of new nodes happens frequently.

Using the notion of a presence function, many interesting concepts from static graph theory can
be generalized to a dynamic version. For example, the neighborhood of a node $\lnode$ at time $\timeidx$ can be defined as the set of all nodes
$\rnode$ with $\nodePres(\rnode, \timeidx) = 1$ for which an edge $(\lnode, \rnode) \in \edges$ exists such that $\edgePres((\lnode, \rnode), \timeidx) = 1$.
Similarly, a path at time $\timeidx$ between two nodes $\lnode$ and $\rnode$ can be defined as a sequence of
edges $(\lnode_1, \rnode_1), \ldots, (\lnode_K, \rnode_K)$, such that $\lnode = \lnode_1$, $\rnode = \rnode_K$
and for all $k$ $(\lnode_k, \rnode_k) \in \edges$, $\lnode_{k+1} = \rnode_k$, $\nodePres(\lnode_k, \timeidx) = 1$, and
$\edgePres((\lnode_k, \rnode_k), \timeidx) = 1$. Two nodes $\lnode$ and $\rnode$
can be called connected at time $\timeidx$ if a path between them exists at time $\timeidx$.
Further, we can define the \emph{temporal subgraph} $\graph_\timeidx$ of graph $\graph$ at time $\timeidx$ as the graph of
all nodes and edges of $\graph$ which are present at time $\timeidx$.

Note that we have assumed discrete time in our definition of a time-varying graph. This is justified
by the following consideration: If a graph is embedded in continuous time, changes to the graph
take the form of value changes in the node or edge presence function, which happen discretely,
because a presence function can only take the values $0$ or $1$. We call points in continuous time
where such a discrete change occurs \emph{events}. Assuming that there are only
finitely many such events, we can write all events in the lifetime of a graph as an ascending sequence
$\timeidx_1, \ldots, \timeidx_K$. Accordingly, all changes in the graph are fully described by the
sequence of temporal subgraphs $\graph_{\timeidx_1}, \ldots, \graph_{\timeidx_K}$ \cite{Casteigts2012,Scherrer2008}.
Therefore, even time-varying graphs defined on continuous time can be fully described by considering
the discrete lifetime $\{1, \ldots, K\}$.

\paragraph{Sequential Dynamical Systems:} Sequential dynamical systems (SDS) have been introduced by
Barret, Reidys and Mortvart to generalize cellular automata to arbitrary graphical structures
\cite{Barrett2000,Barrett1999}. In essence, they assign a binary state to each node $\node$ in
a static graph $\graph = (\nodes, \edges)$. This state is updated according to a function
$\fun_\node$ which maps the current states of the node itself and all of its neighbors to the next state of the node $\node$.
This induces a discrete dynamical system on graphs (where edges and neighborhoods stay fixed) \cite{Barrett2000,Barrett2003,Barrett1999}.
Interestingly, SDSs can be related to time-varying graphs by interpreting the binary state of a node $\node$ at
time $\timeidx$ as the value of its presence function $\nodePres(\node, \timeidx)$.
As such, sequential dynamical systems provide a compact model for the node presence function of a
time-varying graph. Furthermore, if the graph dynamics is governed by a known SDS, its future
can be predicted by simply simulating the dynamic system. Unfortunately, if the underlying SDS is
unknown, this technique can not be applied, and to our knowledge, no learning schemes exists to date
to infer an underlying SDS from data. Therefore, other predictive methods are required.

\subsection{Predicting Changes in Graphs}

In accordance with classical time series prediction, one can describe time series prediction in
graphs as predicting the next temporal subgraph $\graph_{\timeidx+1}$ given a sequence of temporal
subgraphs $\graph_0, \ldots, \graph_\timeidx$ in a time-varying graph. To our knowledge, there exists
no approach which addresses this problem in this general form. However, more specific subproblems
have been addressed in the literature.

\paragraph{Link Prediction:} In the realm of social network analysis, Liben-Nowell and Kleinberg have formulated
the \emph{link prediction problem}, which can be stated as: Given a sequence of temporal subgraphs
$\graph_0, \ldots, \graph_\timeidx$ for a time-varying graph $\graph$, which edges will be added to
the graph in the next time step, i.e.\ for which edges do we find $\edgePres(\edge, \timeidx) = 0$
but $\edgePres(\edge, \timeidx + 1) = 1$ \cite{Nowell2007,Yang2015}? For example, given all past collaborations
in a scientific community, can we predict new collaborations in the future?
The simplest approach to address this challenge is to compute a similarity index between nodes,
rank all non-existing edges $(\lnode, \rnode)$ according to the similarity index of their nodes $\simi(\lnode, \rnode)$
and predict all edges which have an index above a certain threshold \cite{Nowell2007,Lichtenwalter2010}.
Typical similarity indices to this end include the number of common neighbors at time $\timeidx$,
the Jaccard index at time $\timeidx$ or the Adar index at time $\timeidx$ \cite{Nowell2007}.
A more recent approach is to train a classifier that predicts the value of the edge presence
function $\edgePres(\edge, \timeidx+1)$ for all edges with $\edgePres(\edge, \timeidx) = 0$
using a vectorial feature representation of the edge $\edge$ at time $\timeidx$, where features
include the similarity indices discussed above \cite{Lichtenwalter2010}. In a survey, Lü and
Zhou further list maximum-likelihood approaches on stochastic models and probabilistic relational
models for link prediction \cite{Lue2011}.

\paragraph{Growth models:} In a seminal paper, Barabási and Albert described a simple model to incrementally
grow an undirected graph node by node from a small, fully connected seed graph \cite{Barabasi1999}. Since then,
many other models of graph growth have emerged, most notably stochastic block models and latent
space models \cite{Clauset2013,Goldenberg2010}. Stochastic block models assign each node to a block
and model the probability of an edge between two nodes only dependent on their
respective blocks \cite{Holland1983}. Latent space models embed all nodes in an underlying, latent
space and model the probability of an edge depending on the distance in this space \cite{Hoff2002}.
Both classes of models can be used for link prediction as well as graph generation. Further, they can be
trained with pre-observed data in order to provide more accurate models of the data. However, none of
the discussed models addresses the question of time series prediction in general because the deletion of
nodes or edges or label changes are not covered by the models.

\subsection{Graph Dissimilarities and Kernels}

Instead of predicting the next graph in a sequence directly, one can consider indirect approaches,
which base a prediction on pairwise dissimilarities $\diss(\graph, \graph')$ or kernels
$\kernel(\graph, \graph')$. As a simple example, consider a 1-nearest-neighbor approach:
We first aggregate a database of training data consisting of graphs $\graph_\timeidx$ and their
respective successors $\graph_{\timeidx+1}$. If we are confronted with a new graph $\graph'$, we
simply predict the graph $\graph_{\timeidx+1}$ such that $\diss(\graph_\timeidx, \graph')$ is
minimized or $\kernel(\graph_\timeidx, \graph')$ is maximized.
The task of quantifying distance and similarity between graphs has attracted considerable research
efforts, which we can divide roughly into two main streams: graph edit distances and graph kernels.

\paragraph{Graph Edit Distance:} The graph edit distance between two graphs $\graph$ and $\graph'$
is traditionally defined as the minimum number of edit operations required to transform $\graph$
into $\graph'$. Permitted edit operations include node insertion, node deletion, edge insertion,
edge deletion, and the substitution of labels in nodes or edges \cite{Sanfeliu1983}. This problem
is a generalization of the classic string or tree edit distance, which is defined as the minimum number
of operations required to transform a string into another or a tree into another respectively
\cite{Levenshtein1965,Zhang1989}. Unfortunately, while the string edit distance and the tree edit
distance can be efficiently computed in $\effic(n^2)$ and $\effic(n^4)$ respectively, computing
the exact graph edit distance is NP-hard \cite{Zeng2009}. However, many approximation schemes
exist, e.g.\ relying on self-organizing maps, Gaussian mixture models, graph kernels or binary
linear programming \cite{Gao2010}. A particularly simple approximation scheme is to order the
nodes of a graph in a sequence and then apply a standard string edit distance measure on these
sequences (refer e.g.\ to \cite{Paassen2016,RoblesKelly2003,RoblesKelly2005}). For our experiments
on real-world Java data we will rely on a kernel over such an approximated graph distance as suggested
in \cite{Paassen2016}.

Note that for labeled graphs one may desire to assign non-uniform costs for substituting labels.
Indeed, some labels may be semantically more related than others, implying a lower edit cost.
Learning such edit costs from data is the topic of structure metric learning, which has mainly been
investigated for string edit distances \cite{Bellet2013}. However, if string edit distance is applied
as a substitute for graph edits, the results apply to graphs as well \cite{Paassen2016}.

\paragraph{Graph Kernels:} Complementary to the view of graph edit distances, graph kernels
represent the similarity between two graphs instead of their dissimilarity. The only formal requirement
on a graph kernel is that it implicitly or explicitly maps a graph $\graph$ to a vectorial
feature representation $\kernelfeat(\graph)$ and computes the pairwise kernel values of two graphs
$\graph$ and $\graph'$ as the dot-product of their feature vectors, that is, $\kernel(\graph, \graph') =
\transp{\kernelfeat(\graph)} \cdot \kernelfeat(\graph')$ \cite{DaSanMartino2010}. If each graph has a
unique representation (that is, $\kernelfeat$ is injective) computing such
a kernel is at least as hard as the graph isomorphy problem \cite{Gaertner2003}. Thus, efficient
graph kernels rely on a non-injective feature embedding, which is still expressive enough to capture
important differences between graphs \cite{DaSanMartino2010}. A particularly popular class
of graph kernels are walk kernels which decompose the kernel between two graphs into kernels
between paths which can be taken in the graphs \cite{Borgwardt2005,DaSanMartino2010,Feragen2013}.
More recently, advances have been made using decompositions of graphs into constituent parts
instead of parts \cite{da_san_martino_ordered_2016,shervashidze_weisfeiler-lehman_2011}. 
The overall kernel is constructed via a sum of kernel values between the parts, which is guaranteed
to preserve the kernel property according to the convolutional kernel principle \cite{Haussler1999}.
In our experiments on artificial data, we apply the shortest-path-length kernel suggested by Borgwardt and colleagues,
which compares the lengths of shortest paths in both graphs to construct an overall graph kernel,
which is sufficiently expressive for these simple cases \cite{Borgwardt2005}.

Any of these quantitative measures of distance or similarity supports time series prediction in a
1-nearest neighbor fashion as stated above. However, we would like to apply more sophisticated
prediction methods as well. To this end, we turn to kernel-based methods.

\subsection{Kernel-Based approaches for Vectorial Time Series Prediction}

The idea to apply kernel-based methods for time series prediction as such is not new.
One popular example is the use of support vector regression with wide-ranging
applications in finance, business, environmental research and engineering \cite{Sapankevych2009}.
Another example are gaussian processes to predict chemical processes \cite{Girard2003}, motion data \cite{Wang2005}
and physics data \cite{Roberts2012}. If a graph kernel is used that provides an explicit vectorial
embedding of graphs, then these methods can be readily applied for time series prediction on graph
data. However, in this work, we generalize this approach to kernels where only an \emph{implicit}
vectorial embedding is available. This enables us to use a broader range of graph distances and
kernels, such as the graph edit distance approximation for our Java program data.

\section{Time Series Prediction for Graphs}
\label{sec:theory}

\subsection{Time Series Prediction as a Regression Problem}

Relying on the notation of time-varying graphs as introduced above, we can describe a time series
of graphs as a sequence of temporal subgraphs $\graph_0, \ldots, \graph_\timeidx$. Time series
prediction is the problem of predicting the next graph in the series, $\graph_{\timeidx+1}$. We first
transform this problem into a regression problem as suggested by Sapankevych and colleagues, that is,
we try to learn a function $\fun$ which maps the past $K$ states of a time series to a successor state
\cite{Sapankevych2009}.
Relying on distance- and kernel-based approaches we can address this problem by computing pairwise
distance- or kernel values on graph sequences of length $K$, that is, $\diss((\graph_{\timeidx-K}, \ldots, \graph_\timeidx), (\graph'_{\timeidx'-K}, \ldots, {\graph'}_{\timeidx'}))$
or $\kernel((\graph_{\timeidx-K}, \ldots, \graph_\timeidx), (\graph'_{\timeidx'-K}, \ldots, {\graph'}_{\timeidx'}))$
respectively. Using the sequential edit distance framework \cite{Paassen2016} or the convolutional kernel principle
\cite{DaSanMartino2010,Gaertner2003}, one can easily construct distances and kernels on graph sequences
from distances and kernels from graphs. However, for simplicity, we will assume $K=1$ in the remainder
of this paper, such that simple graph distances or kernels are sufficient. This is equivalent
to the Markov assumption, that is, we assume that the next temporal subgraph $\graph_{\timeidx+1}$
only depends on the temporal subgraph $\graph_\timeidx$ and is conditionally independent from all
temporal subgraphs $\graph_0, \ldots, \graph_{\timeidx-1}$. Note that the Markov assumption is quite
common in the related literature. For example, sequential dynamical systems make the very same
assumption \cite{Barrett2000}, as do most link prediction methods \cite{Lue2011} and
graph growth models \cite{Clauset2013,Goldenberg2010}. Further note that our framework generalizes
to the non-Markovian case by replacing the graph distance or kernel to a sequence-of-graph
distance or kernel and using everything else as-is.

Under the Markov assumption we can express our training data as tuples
$\{ (\point_\dataidx, \outpoint_\dataidx) \}$ for $\dataidx \in \{1, \ldots, \datalim\}$
where $\outpoint_\dataidx$ is the successor of $\point_\dataidx$ in some time series, that is,
$\outpoint_\dataidx = \graph_{\timeidx+1}$ and $\point_\dataidx = \graph_\timeidx$. We denote
the unseen test data point, for which a prediction is desired, as $\point$.

We proceed by introducing three non-parametric regression techniques, namely $1$-nearest neighbor,
kernel regression and Gaussian process regression. We then revisit some basic research on dissimilarities,
similarities and kernels and leverage these insights to derive time series prediction purely in
latent spaces without referring to any explicit vectorial form. Finally, we discuss how to speed
up Gaussian process prediction from cubic to linear time.

\subsection{Non-Parametric Regression techniques}

In this section, we introduce three non-parametric regression techniques in their standard form,
assuming vectorial input and output data. Further below we explain how these methods can be
applied to structured input and output data. We assume that a training data set of
tuples $\{ (\vec \point_\dataidx, \vec \outpoint_\dataidx) \}$ is available, where $\vec \outpoint_\dataidx$
is the desired output for input $\vec \point_\dataidx$. Further, we assume that a dissimilarity
$\diss$ or a kernel $\kernel$ on the input space is available wherever needed.

\paragraph{1-Nearest Neighbor (1-NN):} We define the predictive function for one-nearest neighbor
regression (1-NN) as follows:
\begin{equation}
	\fun(\vec \point) := \vec \outpoint_{\dataidx^+}  \text{ where }
	\dataidx^+ = \argmin_{\dataidx \in \{1, \ldots, \datalim\}} \diss(\vec \point, \vec \point_\dataidx)
	\label{eq:1nn}
\end{equation}
Note that the predictive function of 1-NN is not smooth, because the $\argmin$ result is ill-defined
if there exist two $\dataidx, \dataidx'$ with $\dataidx \neq \dataidx'$ and $\vec \outpoint_\dataidx \neq \vec \outpoint_{\dataidx'}$
such that $\diss(\vec \point, \vec \point_\dataidx) = \diss(\vec \point, \vec \point_{\dataidx'})$.
At these points, the predictive function is discontinuous and jumps between $\vec \outpoint_\dataidx$ and
$\vec \outpoint_{\dataidx'}$. A simple way to address this issue is kernel regression.

\paragraph{Kernel Regression (KR):} Kernel regression was first proposed by Nadaraya and Watson and can
be seen as a generalization of 1-nearest neighbor to a smooth predictive function $\fun$ using
a kernel $\kernel$ instead of a dissimilarity \cite{Nadaraya1964}. The predictive function is given as:
\begin{equation}
\fun(\vec \point) := \frac{
	\sum_{\dataidx = 1}^{\datalim} \kernel(\vec \point, \vec \point_\dataidx) \cdot \vec \outpoint_\dataidx
}{
	\sum_{\dataidx = 1}^{\datalim} \kernel(\vec \point, \vec \point_\dataidx)
}
\end{equation}
Note that kernel regression generally assumes kernel values to be positive and requires at least one
$\dataidx$ for which $\kernel(\vec \point, \vec \point_\dataidx) > 0$, that is, 
if the test data point is not similar to any training data point, the prediction degenerates.
Another limitation of kernel regression is that an exact reproduction of the training data is not possible,
i.e.\ $\fun(\vec \point_\dataidx) \neq \vec \outpoint_\dataidx$. To achieve both, training data reproduction
as well as a smooth predictive function, we turn to Gaussian process regression.

\paragraph{Gaussian Process Regression (GPR):} In Gaussian process regression (GPR) we assume that
the output points (training as well as test) are a realization of a multivariate random variable with a Gaussian distribution \cite{Rasmussen2005}.
The model extends KR in several ways. First, we can encode prior knowledge regarding the output points
via the mean of our prior distribution, denoted as $\vec \prior_\dataidx$ and $\vec \prior$ for $\vec \outpoint_\dataidx$ and $\vec \outpoint$ respectively.
Second, we can cover Gaussian noise on our training output points within our model. For this noise, we assume mean $0$
and standard deviation $\nDevNoise$.

Let now $\kernel$ be a kernel on $\inspace$. We define
\begin{align}
\kernelvec &:= \big(\kernel(\vec \point, \vec \point_1), \dots, \kernel(\vec \point, \vec \point_\datalim)\big) \text{ and} \\
\kernelmat &:= \big(\kernel(\vec \point_\dataidx, \vec \point_{\dataidx'})\big)_{\dataidx, \dataidx' = 1 \ldots \datalim}
\end{align}
Then under the GP model, the conditional density function of the output points given the input points is
\begin{equation}
\dens( \vec \outpoint_1, \ldots, \vec \outpoint_\datalim, \vec \outpoint | \vec \point_1, \ldots, \vec \point_\datalim, \vec \point) = 
\nDens\Big(\vec \prior_1, \dots, \vec \prior_\datalim, \vec \prior ,
	\begin{pmatrix}
		\kernelmat + \nDevNoise^2 \cdot \eye^\datalim & \transp{\kernelvec} \\
		\kernelvec & \kernel(\vec \point, \vec \point)
	\end{pmatrix}
\Big)
\end{equation}
where $\eye^\datalim$ is the $\datalim$-dimensional identity matrix and $\nDens(\vec \nMean, \nCov)$
is the multivariate Gaussian probability density function for mean $\vec \nMean$ and covariance matrix
$\nCov$. Note that our assumed distribution takes \emph{all} outputs $\vec \outpoint_1, \ldots, \vec \outpoint_\datalim, \vec \outpoint$
as argument, not just a single point. The posterior distribution for just $\vec \outpoint$
can be obtained by marginalization:
\begin{theorem}[Gaussian Process Posterior Distribution]
Let $\outmat$ be the matrix $\transp{(\vec \outpoint_1, \ldots, \vec \outpoint_\datalim)}$ and
$\Prior := \transp{(\vec \prior_1, \ldots, \vec \prior_\datalim)}$.
Then the posterior density function for Gaussian process regression is given as:
\begin{align}
\dens(\vec \outpoint &| \vec \point, \vec \point_1, \dots, \vec \point_\datalim,
	\vec \outpoint_1, \dots, \vec \outpoint_\datalim) =
	\nDens\Big(\vec \nMean , \nDev^2 \cdot \eye^\dims\Big) \quad \mbox{where}\\
\vec \nMean &= \transp{\vec  \prior} +
	\kernelvec \cdot (\kernelmat + \nDevNoise^2 \cdot \eye^\datalim)^{-1}
		\cdot (\outmat - \Prior)\\
\nDev^2 &= \kernel(\vec \point , \vec \point)
	- \kernelvec \cdot (\kernelmat + \nDevNoise^2 \cdot \eye^\datalim)^{-1}
		\cdot \transp{{\kernelvec}} \label{eq:var}
\end{align}
We call $\vec \nMean$ the \emph{predictive mean} and $\nDev^2$ the \emph{predictive variance}.
\begin{proof}
Refer e.g.\ to \cite[p. 27]{Rasmussen2005}. 
\end{proof}
\end{theorem}
Note that the posterior distribution is, again, a Gaussian distribution. For such a distribution,
the mean corresponds to the point of maximum probability density, such that we can define our
predictive function as $\fun(\vec \point) := \transp{\vec \nMean}$ where $\vec \nMean$ is the predictive
mean of the posterior distribution for point $\vec \point$. Further note that the predictive mean
becomes the prior mean if $\kernelvec$ is the zero vector, i.e.\ if the test data point is not similar
to any training data point.

The main drawback of GPs is their high computational complexity: For training, the inversion of the
matrix $(\kernelmat + \nDevNoise^2 \cdot \eye^\datalim)^{-1}$ requires cubic time. We will address
this problem in section~\ref{sec:linear_time}.

\subsection{Dissimilarities, Similarities and Kernels}

Each of the methods described above relies either on a dissimilarity (in the case of 1-NN), or
on a kernel (in the case of KR and GPR). Note that we have not strictly defined these concepts
until now. Indeed, dissimilarities as well as similarities are inherently ill-defined concepts.
A dissimilarity for the set $\inspace$ is any function of the form $\diss : \inspace \times \inspace \to \R$,
such that $\diss$ decreases if the two input arguments are in some sense more related to each other \cite{Pekalska2005}.
Conversely, a similarity for the set $\inspace$ is any function $\simi : \inspace \times \inspace \to \R$
which increases if the two input arguments are in some sense more related to each other \cite{Pekalska2005}.
While in many cases, more rigorous criteria apply as well (such as non-negativity, symmetry,
or that any element is most similar to itself) any or all of them may be violated in practice \cite{Schleif2016}.
On the other hand, kernels are strictly defined as inner products. If we require a kernel to apply
our prediction method but only have a dissimilarity or a similarity available, we require a conversion
method to obtain a corresponding kernel.
In a first step we note that it is simple to convert from a dissimilarity $\diss$ to a similarity $\simi$.
We can simply apply any monotonously decreasing function.
In the context of this work, we will apply the radial basis function transformation
\begin{equation}
	\simi_\diss(\point, \point') := \exp\left( - \frac{1}{2} \frac{\diss(\point, \point')^2}{\rbfband^2} \right)
\end{equation}
where $\rbfband$ is a positive real number called \emph{bandwidth}.

Further, for any finite set of data points $\point_1, \ldots, \point_\datalim$, 
we can transform a symmetric similarity into a kernel. Let $\simimat$ be a symmetric matrix of pairwise similarities
for these data points with the entries $\simimat_{\dataidx \dataidx'} = \simi(\point_\dataidx, \point_{\dataidx'})$.
Such a symmetric similarity matrix $\simimat$ is a kernel matrix if and only if it is positive semi-definite \cite{Pekalska2005}.
Positive semi-definiteness can be enforced by applying an eigenvalue decomposition $\simimat = \mat{U} \mat{\Lambda} \mat{V}$
and removing the negative eigenvalues in the matrix $\mat{\Lambda}$, e.g.\ by setting them to zero or
by taking their absolute value. The resulting matrix $\mat{\tilde \Lambda}$ can then be applied to
obtain a kernel matrix $\kernelmat = \mat{U} \mat{\tilde \Lambda} \mat{V}$
\cite{Gisbrecht2015,Pekalska2005,Schleif2016}. Note that the eigenvalue decomposition of a similarity matrix
has cubic time-complexity in the number of data points. However, linear-time approximations have
recently been discovered based on the Nyström method \cite{Gisbrecht2015}. 
Therefore, in our further work, we can assume that a kernel can be obtained as needed.

\subsection{Prediction in Kernel and Dissimilarity Spaces}

In introducing the three regression techniques above we have assumed that each data point is a vector,
such that algebraic operations like scalar multiplication (for kernel regression), as well as matrix-vector
multiplication and vector addition (for Gaussian process regression) are permitted. In the case of graphs,
such operations are not well-defined. However, we can rely on prior work regarding vectorial embeddings of dissimilarities
and similarities to define our regression in a latent space. In particular:

\begin{theorem}[Pseudo-Euclidean Embeddings \cite{Pekalska2005}]\label{thm:euclidean_emb}
For dissimilarities: Let $\point_1, \ldots, \point_\datalim$ be some points from a set $\inspace$, and let
$\diss : \inspace \times \inspace  \to \R^+$ be a function such that for all
$\point_\dataidx, \point_{\dataidx'}$ it holds: $\diss(\point_\dataidx, \point_{\dataidx'}) \geq 0$,
$\diss(\point_\dataidx, \point_\dataidx) = 0$ and $\diss(\point_\dataidx, \point_{\dataidx'}) = \diss(\point_{\dataidx'}, \point_\dataidx)$.
Then, there exists a vector space $\dissspace \subset \R^\dims$
and a function $\dissfeat : \inspace \to \dissspace$ such that for all
$\dataidx, \dataidx' \in \{1, \ldots, \datalim\}$ it holds:
\begin{equation}
\diss(\point_\dataidx, \point_{\dataidx'}) = \sqrt{
	\transp{\Big(\dissfeat(\point_\dataidx) - \dissfeat(\point_{\dataidx'})\Big)} \cdot
	\Lambda \cdot
	\Big(\dissfeat(\point_\dataidx) - \dissfeat(\point_{\dataidx'})\Big)
}
\end{equation}
for some diagonal matrix $\Lambda \in \{-1, 0, 1\}^{\dims \times \dims}$.

For kernels: Let $\kernel$ be a kernel on $\inspace$. Then there exists an Euclidean space $\kernelspace$
and a function $\kernelfeat : \inspace \to \kernelspace$ such that for all
$\dataidx, \dataidx' \in \{1, \ldots, \datalim\}$ it holds:
\begin{equation}
\kernel(\point_\dataidx, \point_{\dataidx'}) =
	\transp{\kernelfeat(\point_\dataidx)} \cdot \kernelfeat(\point_{\dataidx'})
\end{equation}
\begin{proof}
Refer to \cite{Pekalska2005}.
\end{proof}
\end{theorem}

The key idea of our approach to time series prediction for structured data (i.e.\ graphs) is to apply time series prediction
in the implicit (pseudo-)Euclidean space $\kernelspace$, without any need for an explicit embedding in $\kernelspace$.
This is obvious for 1-NN: Our prediction is just the successor to the closest training data point. This is a graph
itself, providing us with a function mapping directly from graphs as inputs to graphs as outputs. The situation
is less clear for KR or GPR. Here, our approach is to express the predictive output as a linear combination of
known data points, which permits further processing as we will show later. In particular, we can proof that the
resulting linear combinations are convex or at least affine. Linear, convex and affine combinations are defined as follows:
\begin{definition}[Linear, Affine, and Convex Combinations]
Let $\vec \point_1, \ldots, \vec \point_\datalim$ be vectors from some vector space $\inspace$
and let $\coeff_1, \ldots, \coeff_\datalim$ be real numbers. The sum
$\sum_{\dataidx = 1}^\datalim \coeff_\dataidx \cdot \vec \point_\dataidx$
is called a \emph{linear combination} and the numbers $\coeff_\dataidx$ \emph{linear coefficients}.
If it holds $\sum_{\dataidx = 1}^\datalim \coeff_\dataidx = 1$ the linear combination is called
\emph{affine}. If it additionally holds that for all $\dataidx$ $\coeff_\dataidx \geq 0$ the affine
combination is called \emph{convex}.
\end{definition}

Our proof that the prediction provided by GPR is an affine combination requires two additional assumptions.
First, we require a certain, natural prior, namely: In the absence of any further knowledge, our best prediction
for the point $\point$ is the identity, i.e.\ staying where we are.
In this case, we obtain $\vec \prior = \vec \point$ and $\vec \prior_\dataidx = \vec \point_\dataidx$ and the
predictive mean
\begin{equation}
\vec \nMean = \transp{\vec \point} + \kernelvec \cdot (\kernelmat + \nDevNoise^2 \cdot \eye^\datalim)^{-1} \cdot (\outmat - \inmat)
\end{equation}
where $\inmat = \transp{(\vec \point_1, \ldots, \vec \point_\datalim)}$. The predictive variance
is still the same as in equation~\ref{eq:var}. We will assume this prior in the remainder of this paper.

Further, we observe that our training inputs and outputs for time series prediction have a special form: Training data
is presented in the form of sequences $\point^\trajidx_1, \ldots, \point^\trajidx_{\timelim_\trajidx}$
for $\trajidx \in \{1, \ldots, \trajlim\}$ where $\point^\trajidx_{\timeidx+1}$ is the successor of $\point^\trajidx_\timeidx$.
This implies that each point that is a predecessor of another point is also a successor of another point (except for the
first element in each sequence) and vice versa (except for the last element in each sequence).

\begin{theorem}[Predictive results as combinations] \label{thm:affine_combs}
If training data is provided in terms of sequences $\point^\trajidx_1, \ldots, \point^\trajidx_{\timelim_\trajidx}$
for $\trajidx \in \{1, \ldots, \trajlim\}$ where $\point^\trajidx_{\timeidx+1}$ is the successor of $\point^\trajidx_\timeidx$, then it holds:
\begin{enumerate}
	\item The predictive result for 1-NN is a single training data point (i.e.\ a convex combination of only a single data point).
	\item The predictive result for KR is an affine combination of training data points.
	\item If the prior $\prior = \point$ and $\prior^\trajidx_\timeidx = \point^\trajidx_\timeidx$ is assumed, the predictive
result for GPR is an affine combination of training data points and the test data point, where the test data point has the coefficient $1$.
\end{enumerate}
\begin{proof}
\begin{enumerate}
	\item This follows directly from the form of the predictive function in equation~\ref{eq:1nn}
	\item The coefficients are given as
	\begin{equation}
	\frac{
		\kernel(\point, \point^\trajidx_\timeidx)
	}{
		\sum_{\trajidx'=1}^\trajlim \sum_{\timeidx' = 1}^{\timelim_{\trajidx'}-1} \kernel(\point, \point^{\trajidx'}_{\timeidx'})
	}
	\end{equation}
	for any point $\point^\trajidx_{\timeidx+1}$ with $\trajidx \in \{1, \ldots, \trajlim\}$ and
	$\timeidx \in \{1, \ldots, \timelim_\trajidx - 1\}$. This combination is affine due to the normalization by $\sum_{\trajidx'=1}^\trajlim \sum_{\timeidx' = 1}^{\timelim_{\trajidx'}-1} \kernel(\point, \point^{\trajidx'}_{\timeidx'})$.
	If the kernel is non-negative, the combination is convex.
	\item We define $\vec \gpcoeff = (\gpcoeff^1_1, \ldots, \gpcoeff^1_{\timelim_1 - 1}, \ldots,
\gpcoeff^\trajlim_1, \ldots, \gpcoeff^\trajlim_{\timelim_\trajlim -1}) := \kernelvec \cdot 
(\kernelmat + \nDevNoise^2 \cdot \eye^\datalim)^{-1}$. 
The predictive mean is given as $\vec \nMean = \vec \point + \vec \gpcoeff 
\cdot (\outmat - \inmat)$ which yields
\begin{align}
	\vec \nMean &= \vec \point +
	 \sum_{\trajidx = 1}^\trajlim \sum_{\timeidx = 1}^{\timelim_\trajidx - 1}
		\gpcoeff^\trajidx_\timeidx \cdot \Big(
			\vec \point^\trajidx_{\timeidx+1} -
			\vec \point^\trajidx_\timeidx
			\Big)\\
	&= \vec \point +
	\sum_{\trajidx = 1}^\trajlim
		- \gpcoeff^\trajidx_1 \cdot \vec \point^\trajidx_1 +
		\bigg(\sum_{\timeidx = 2}^{\timelim_\trajidx - 1}
			(\gpcoeff^\trajidx_{\timeidx - 1} - \gpcoeff^\trajidx_\timeidx) \cdot
				 \vec \point^\trajidx_\timeidx
		\bigg) + \gpcoeff^\trajidx_{\timelim_\trajidx - 1} \cdot
			\vec \point^\trajidx_{\timelim_\trajidx}
\end{align}
Thus, the coefficients within each sequence are $-\gpcoeff^\trajidx_1$, $\gpcoeff^\trajidx_1 - \gpcoeff^\trajidx_2$, $\ldots$,
$\gpcoeff^\trajidx_{\timelim_\trajidx-2} - \gpcoeff^\trajidx_{\timelim_\trajidx-1}$, $\gpcoeff^\trajidx_{\timelim_\trajidx-1}$.
These coefficients add up to zero. Finally, we have a coefficient of $1$ for $\vec \point$, which is our prior for the
prediction. Therefore, the overall sum of all coefficients is $1$ and the combination is affine.
\end{enumerate}
\end{proof}
\end{theorem}

In effect, we can reformulate the predictive function as $\fun : \inspace \to \R^{\datalim+1}$ mapping
a test data point to the coefficients of an affine combination, which represent our actual predictive
result in the (pseudo-)Euclidean space $\kernelspace$. Note that these coefficients do not provide us with a primal space
representation of our data, i.e.\ we do not know what the graph which corresponds to a
particular affine combination looks like. Indeed, as graph kernels are generally not injective \cite{Gaertner2003}, there might be multiple
graphs that correspond to the same affine combination. Finding such a graph is called the \emph{kernel pre-image problem}
and is hard to solve even for vectorial data \cite{bakir_learning_2003,bakir_learning_2004,Kwok2004}. This poses a challenge with respect
to further processing: How do we interpret a data point for which we have no explicit representation \cite{Hofmann2014}?

Fortunately, we can still address many classical questions of data analysis for such a representation
relying on our already existing implicit embedding in $\kernelspace$. We can simply extend this
embedding for the predicted point via the affine combination as follows:
\begin{theorem}[Distances and Kernels for Affine Combinations]
For dissimilarities: Let $\point_1, \ldots, \point_\datalim$, $\dissfeat$, $\diss$ and $\Lambda$ be as in theorem~\ref{thm:euclidean_emb}.
Let $\dissmat^2$ be the matrix of squared pairwise dissimilarities between the points $\point_1, \ldots, \point_\datalim$,
and let $\vec \coeff$ be a $1 \times \datalim$ vector of affine coefficients. Then it holds for any $\dataidx \in \{1, \ldots, \datalim\}$:
\begin{equation}
\transp{\Big(\dissfeat(\point_\dataidx) - \sum_{\dataidx' = 1}^\datalim \coeff_{\dataidx'} \dissfeat(\point_{\dataidx'})\Big)} \cdot \Lambda \cdot
	\Big(\dissfeat(\point_\dataidx) - \sum_{\dataidx' = 1}^\datalim \coeff_{\dataidx'} \dissfeat(\point_{\dataidx'})\Big)
= \sum_{\dataidx' = 1}^\datalim \coeff_{\dataidx'} \diss(\point_\dataidx, \point_{\dataidx'})^2
- \inv{2} \vec \coeff \cdot \dissmat^2 \cdot \transp{\vec \coeff} \label{eq:affin_dist}
\end{equation}

For kernels: Let $\point_1, \ldots, \point_\datalim$, $\kernelfeat$ and $\kernel$ be as in theorem~\ref{thm:euclidean_emb}
and let $\kernelmat$ be the matrix of pairwise kernel values between the points
$\point_1, \ldots, \point_\datalim$. Let $\vec \coeff$ be a $1 \times \datalim$ vector of
linear coefficients. Then it holds for any $\dataidx \in \{1, \ldots, \datalim\}$:
\begin{equation}
\transp{\kernelfeat(\point_\dataidx)} \cdot \left(
	\sum_{\dataidx' = 1}^\datalim \coeff_{\dataidx'} \kernelfeat(\point_{\dataidx'})
\right) =
\sum_{\dataidx' = 1}^\datalim \coeff_{\dataidx'} \kernel(\point_\dataidx, \point_{\dataidx'}) \label{eq:linear_kernel}
\end{equation}
\begin{proof}
Refer to \cite{Hammer2010} for a proof of \ref{eq:affin_dist}. \ref{eq:linear_kernel} follows by simple
linear algebra.
\end{proof}
\end{theorem}
Using this extended embedding, we can simply apply any dissimilarity- or kernel-based method on
the predicted point. For dissimilarities, this includes relational learning vector quantization
for classification \cite{Hammer2014} or relational neural gas for clustering \cite{Hammer2010}.
For kernels, this includes the non-parametric regression techniques discussed here, but also
techniques like kernel vector quantization \cite{Hofmann2012} or support vector machines for classification \cite{Cortes1995}
and kernel variants of $k$-means, SOM and neural gas for clustering \cite{Filippone2008}.
Therefore, we have achieved a full methodological pipeline for preprocessing, prediction and post-processing,
which can be summarized as follows:
\begin{enumerate}
	\item If we intend to use a dissimilarity measure on graphs, we start off by computing
	the matrix of pairwise dissimilarities $\dissmat$ on our training data. If required we
	symmetrize this matrix by setting $\dissmat \gets \frac{1}{2} \cdot (\dissmat + \transp{\dissmat})$
	and set the diagonal to zero.
	Implicitly, this step embeds our training data in a pseudo-Euclidean space $\dissspace$ where $\dissmat$
	are pairwise pseudo-Euclidean distances. We transform this matrix into a similarity matrix $\simimat$ using,
	for example, the radial basis function transformation.
	\item If we intend to use a similarity measure on graphs, we start off by computing the matrix
	of pairwise similarities $\simimat$ on our training data. Otherwise we use the transformed
	matrix from the previous step.
	\item We transform $\simimat$ to a kernel matrix $\kernelmat$ via Eigenvalue correction
	\cite{Gisbrecht2015,Hofmann2014,Pekalska2005}.
	Implicitly, this step embeds our data in an Euclidean space $\kernelspace$ where our
	pairwise kernel values represent inner products of our training data.
	\item For any test data point $\point$ we compute the vector of kernel values $\kernelvec$
	to all training data points.
	\item We apply 1-NN, KR or GPR as needed to infer a prediction for our test data point $\point$
	in form of an affine coefficient vector $\vec \coeff$.
	\item We extend our dissimilarity matrix and/or kernel matrix for the predicted point via
	equations~\ref{eq:affin_dist} and/or \ref{eq:linear_kernel} respectively.
	\item We apply any further dissimilarity- or kernel-based method on the predicted point as
	desired.
\end{enumerate}

The only challenge left is to speed up predictions in GPR to reduce the cubic time
complexity to linear time complexity.

\subsection{Linear Time Predictions}
\label{sec:linear_time}

GP regression involves the inversion of the matrix $(\kernelmat + \nDevNoise^2 \cdot 
\eye^\datalim)$, resulting in $\effic(\datalim^3)$ complexity.
A variety of efficient approximation schemes exist \cite{Rasmussen2005}. Recently, the robust 
Bayesian Committee Machine (rBCM) has been introduced as a particularly fast and accurate 
approximation \cite{Deisenroth2015}. 
The rBCM approach is to distribute the examples into $\clustlim$ disjoint sets, based e.g.\ on 
clustering in the input data space. For each of these sets, a separate GP regression is used, 
yielding the predictive distributions $\nDens(\vec \nMean_\clustidx, \nDev^2_\clustidx)$
for $\clustidx \in \{1, \ldots, \clustlim\}$.
These distributions are combined to the final predictive distribution %
$\nDens(\vec \nMean_\rbcm , \nDev^2_\rbcm )$ with
\begin{align}
\nDev^{-2}_\rbcm  &= \sum_{\clustidx = 1}^\clustlim
				\frac{\clustweight_\clustidx}{\nDev_\clustidx^{2}} +
			\bigg(1 - \sum_{\clustidx = 1}^\clustlim \clustweight_\clustidx \bigg) \cdot
				\inv{\nDev_\mathrm{prior}^2} \label{eq:rbcm_var}\\
\vec \nMean_\rbcm  &= \nDev^2_\rbcm \cdot \Bigg(
	\sum_{\clustidx = 1}^\clustlim
		\frac{\clustweight_\clustidx}{\nDev_\clustidx^{2}} \cdot \vec \nMean_\clustidx
	+ \bigg(1 - \sum_{\clustidx = 1}^\clustlim \clustweight_\clustidx \bigg) \cdot
			\inv{\nDev_\mathrm{prior}^2} \cdot \vec \prior
	\Bigg) \label{eq:rbcm_mean}
\end{align}
Here, $\nDev_\mathrm{prior}^2$ is the variance of the prior for the prediction, which is a new 
meta-parameter introduced in the model. The weights $\clustweight_\clustidx$ can be seen as a 
measure for the predictive power of the single GP experts. As suggested by the authors, we 
use the differential entropy, given as $\clustweight_\clustidx = \inv{2} \cdot \Big( 
\log(\nDev_\mathrm{prior}^2) - \log(\nDev^2_\clustidx) \Big)$ \cite{Deisenroth2015}. Note that
$\vec \prior = \vec \point$ in our case.
This approach results in linear-time complexity if the size of any single cluster is considered
to be constant (i.e.\ the number of clusters is proportional to $\datalim$), such that only a matrix
of constant size has to be inverted.

Two challenges remain to apply rBCM productively within our proposed pipe\-line. First, it remains
to show that the predictive mean of rBCM still has the form of an affine combination. Second, we require
a dissimilarity- or kernel-based clustering scheme which runs in linear time, as to ensure overall
linear-time complexity. Regarding the first issue we show:
\begin{theorem}[rBCM prediction as affine combination]
If training data is provided in terms of sequences $\point^\trajidx_1, \ldots, \point^\trajidx_{\timelim_\trajidx}$
for $\trajidx \in \{1, \ldots, \trajlim\}$ where $\point^\trajidx_{\timeidx+1}$ is the successor of $\point^\trajidx_\timeidx$,
the predictive mean of an rBCM is an affine combination of training data points and the test
data point.
\begin{proof}
We define $\coeff_\clustidx := \frac{\clustweight_\clustidx}{\nDev_\clustidx^{2}}$ and
$\coeff_\mathrm{prior} := \bigg(1 - \sum_{\clustidx = 1}^\clustlim \clustweight_\clustidx \bigg) \cdot
\inv{\nDev_\mathrm{prior}^2}$. For every cluster, the respective predictive mean has the shape
$\vec \nMean_\clustidx = \vec \point + \sum_{\dataidx = 1}^\datalim \gpcoeff^\clustidx_\dataidx \cdot \vec \point_\dataidx$
where $\sum_{\dataidx = 1}^\datalim \gpcoeff^\clustidx_\dataidx = 0$ (as demonstrated in the proof to
theorem~\ref{thm:affine_combs}). So the overall coefficient assigned to $\vec \point$ is
\begin{equation}
	\nDev^2_\rbcm \cdot \left(\sum_{\clustidx = 1}^\clustlim \coeff_\clustidx + \coeff_\mathrm{prior}\right) = 1
\end{equation}
and all other coefficients add up to
\begin{equation}
\sum_{\clustidx = 1}^\clustlim \nDev^2_\rbcm \cdot \coeff_\clustidx \cdot
	\left(\sum_{\dataidx = 1}^\datalim \gpcoeff^\clustidx_\dataidx \right)
= \sum_{\clustidx = 1}^\clustlim \nDev^2_\rbcm \cdot \coeff_\clustidx \cdot 0 = 0
\end{equation}
Therefore, we obtain an affine combination.
\end{proof}
\end{theorem}

\begin{figure}
\begin{center}
\includegraphics[width=0.7\textwidth]{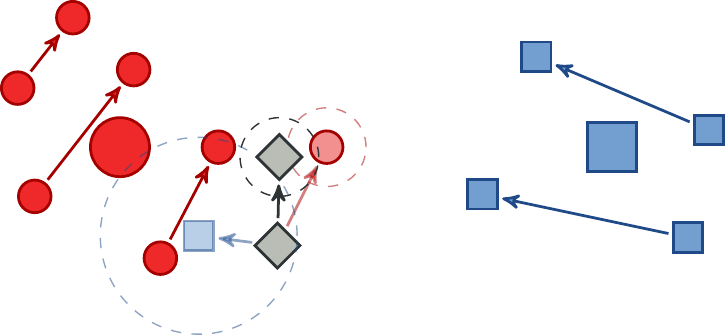}
\end{center}
\caption{An illustration of relational neural gas (RNG) clustering and robust Bayesian committee machine (rBCM) prediction.}
\label{fig:rng_rbcm}
\end{figure}

With regards to the second issue, we apply relational neural gas (RNG) \cite{Hammer2010}.
RNG clusters data points using a Voronoi-tesselation with respect to prototypes. To ensure roughly constant-sized
clusters we have to supply the method with a number of prototypes proportional to the data set size.
Further, we have to ensure that the clustering itself takes only linear time. As such, relational
neural gas requires quadratic time for training. However, it can be sped up by training the prototype
positions only on a constant-sized subset of the data and assigning all remaining data points to the
closest prototype. Computing the distance of all data points to all prototypes requires only the
pairwise distances of all data points to the data points in the training subset, which is in $\effic(\datalim)$.

The approach is illustrated in figure~\ref{fig:rng_rbcm}.
Relational neural gas places prototypes (large circle and square) into the data (small circles and squares) and thereby
distributes data points into disjoint clusters (distinguished by shape). Successor relations are depicted as arrows.
For each cluster, a separate Gaussian process is trained. For a test data point (diamond shape),
each of the GPs provides a separate predictive Gaussian distribution, which are given in terms
of their means (half-tansparent circle and square) and their variance
(dashed, half-transparent circles). The predictive distributions are merged to an overall predictive
distribution with the mean from equation~\ref{eq:rbcm_mean} (solid diamond shape)
and the variance from equation~\ref{eq:rbcm_var} (dashed circle). Note that the overall predictive
distribution is more similar to the prediction of the circle-cluster because the test data point
is closer to this cluster and thus the predictive variance for the circle-cluster is lower, giving
it a higher weight in the merge process.

\section{Experiments}

In our experimental evaluation, we apply the pipeline introduced in the previous section to four
data sets, two of which are theoretical models and two of which are
real-world data sets of Java programs. In all cases, we evaluate the root mean square error (RMSE) of
the prediction for each method in a leave-one-out-crossvalidation over the sequences in our data set.
We denote the current test trajectory as $\point'_1, \ldots, \point'_\timelim$, the training trajectories
as $\{\point^\trajidx_1, \ldots, \point^\trajidx_{\timelim_\trajidx}\}_{\trajidx = 1, \ldots, \trajlim}$,
the predicted affine coefficients for point $\point'_{\timeidx'}$ as $\vec \coeff_{\timeidx'}
= (\coeff^1_{\timeidx', 1}, \ldots, \coeff^\trajlim_{\timeidx', \timelim_\trajlim}, \coeff'_{\timeidx'})$
and the matrix of squared pairwise dissimilarities (including the test data points) as $\dissmat^2$.
Accordingly, the RMSE for each fold has the following form (refer to equation~\ref{eq:affin_dist}).
\begin{equation}
\rmse = \sqrt{\inv{\timelim-1} \sum_{\timeidx'=1}^{\timelim-1}
	\sum_{\trajidx=1}^{\trajlim} \sum_{\timeidx=1}^{\timelim_\trajidx}
	\coeff^\trajidx_{\timeidx', \timeidx} \diss(\point^\trajidx_\timeidx, \point'_{\timeidx'+1})^2
	+ \coeff'_{\timeidx'} \diss(\point'_{\timeidx'}, \point'_{\timeidx'+1})^2
	- \inv{2} \transp{\vec \coeff_{\timeidx'}} \dissmat^2 \vec \coeff_{\timeidx'}}
\end{equation}

We evaluate our four regression models, namely 1-nearest neighbor (1-NN), kernel regression (KR),
Gaussian process regression (GPR) and the robust Bayesian committee machine (rBCM),
as well as the identity function as baseline, i.e.\ we predict the current point as next point.
We optimized the hyper parameters for all methods (i.e.\ the radial basis function bandwidth
$\rbfband$ and the noise standard deviation $\nDevNoise$ for GPR and rBCM) using a random search
with $10$ random trials\footnote{Our implementation of time series prediction is available online
at \url{http://doi.org/10.4119/unibi/2913104}}.
In each trial, we evaluated the RMSE in a nested leave-one-out-crossvalidation over the training
sequences and chose the parameters which corresponded to the lowest RMSE. Let $\avgDiss$ be
the average dissimilarity over the data set. We drew $\rbfband$ from a uniform distribution in
the range $[0.05 \cdot \avgDiss, \avgDiss]$ for the theoretical data sets and fixed it to
$0.3\cdot\avgDiss$ for the Java data sets to avoid the need for a new eigenvalue correction
in each random trial. We drew $\nDevNoise$ from an exponential distribution in the range
$[10^{-3} \cdot \avgDiss, \avgDiss]$ for the theoretical and $[10^{-2} \cdot \avgDiss, \avgDiss]$
for the Java data sets. We fixed the prior standard deviation $\nDev_\mathrm{prior} = \avgDiss$
for all data sets. For rBCM we preprocessed the data via relational neural gas (RNG) clustering
with $\left \lfloor \frac{\datalim}{100} \right \rfloor$ clusters for all data sets.
As this pre-processing could be applied before hyper-parameter selection the runtime overhead of
clustering was negligible and we did not need to rely on the linear-time speedup described above
but could compute the clustering on the whole training data set.

Our experimental hypotheses are that all prediction methods should yield lower RMSE compared
to the baseline (H1), that rBCM should outperform 1-NN and KR (H2) and that rBCM should
be not significantly worse compared to GPR (H3). To evaluate significance we use a
Wilcoxon signed rank sum test.

\subsection{Theoretical Data Sets}

We investigate the following theoretical data sets:

\paragraph{Barabási-Albert model:} This is a simple stochastic model of graph growth in
undirected graphs with hyper-parameters $m_0$, $k$ and $m$ \cite{Barabasi1999}.
The growth process starts with a fully connected initial graph of $m_0$ nodes and adds $m-m_0$
nodes one by one.
Each newly added node is connected to $k$ of the existing nodes which are randomly selected with the
probability $\prob(\lnode) = \deg_\timeidx(\lnode) / ( \sum_\rnode \deg_\timeidx(\rnode) )$
where $\deg_\timeidx$ is the node degree at time
$\timeidx$, i.e.\ $\deg_\timeidx(\rnode) = \sum_\lnode \edgePres((\lnode, \rnode), \timeidx)$.
It has been shown that the edge distribution resulting from this growth model is scale-free, more
specifically the probability of a certain degree $k$ is $\prob(k) \propto k^{-3}$
\cite{Barabasi1999}. Our data set consists of $20$ graphs with
$m = 27$ nodes each, grown from an initial graph of size $m_0 = 3$ and $k = 2$ new edges per node.
This resulted in $500$ graphs overall.

\begin{figure}
\includegraphics[width=\textwidth]{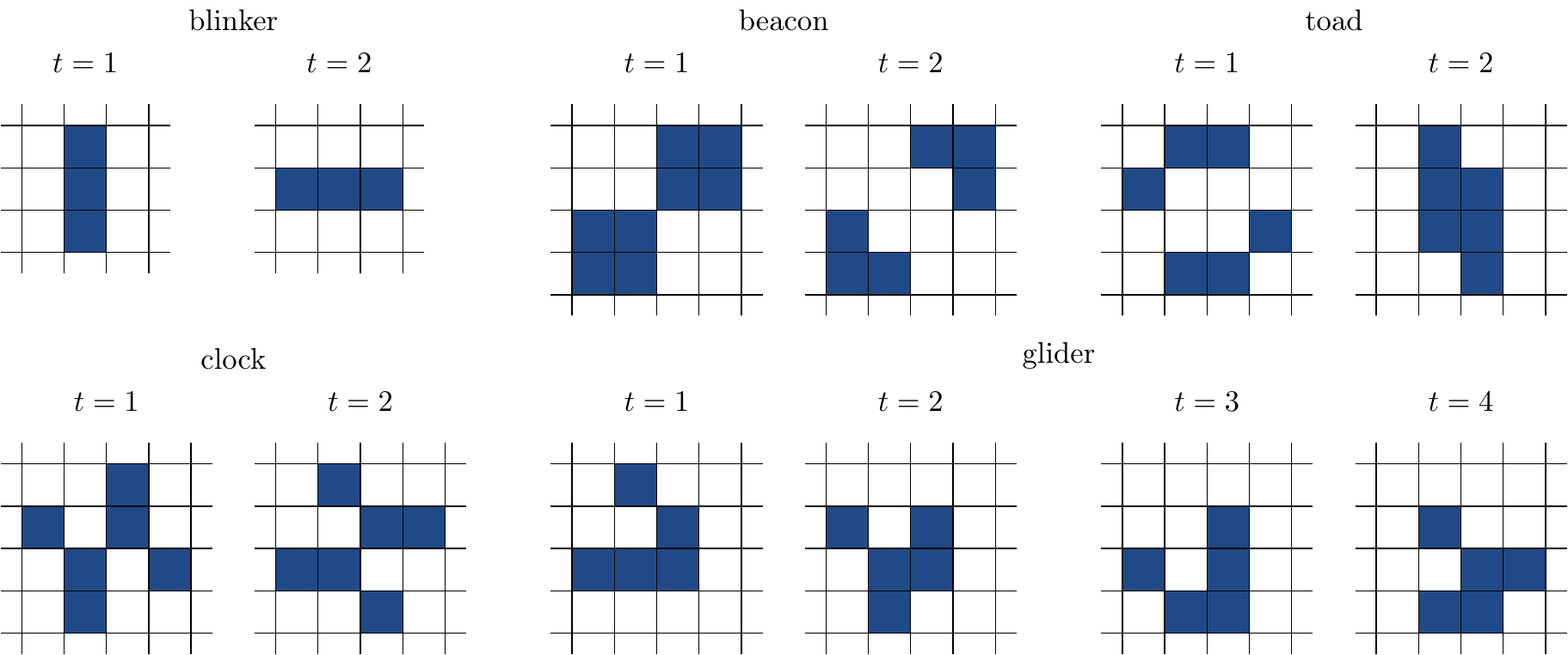}
\caption{The standard patterns used for the \emph{Game of Life}-data set, except for the
\emph{block and glider} pattern. All unique states of the patterns are shown. Note that the state
of glider at $\timeidx = 3$ equals the state at $\timeidx=1$ up to rotation.}
\label{fig:game_of_life}
\end{figure}

\paragraph{Conway's \emph{Game of Life}:} John Conway's \emph{Game of Life} \cite{Gardner1970} is a simple,
2-dimensional cellular automaton model. Nodes are ordered in a regular, 2-dimensional grid
and connected to their eight neighbors in the grid. Let $\neigh(\node)$ denote this eight-
neighborhood in the grid. Then we can describe Conway's \emph{Game of Life} with the following
equations for the node presence function $\nodePres$ and the edge presence function $\edgePres$
respectively:
\begin{align}
\nodePres(\rnode, \timeidx) &=
\begin{cases}
1 & \text{if } 5 \leq \nodePres(\rnode, \timeidx-1) + 2\cdot\sum_{\lnode \in \neigh(\rnode)} \nodePres(\lnode, \timeidx-1) \leq 7 \\
0 & \text{otherwise}
\end{cases} \\
\edgePres((\lnode, \rnode), \timeidx) &=
\begin{cases}
1 & \text{if } \nodePres(\lnode, \timeidx) = 1 \wedge \nodePres(\rnode, \timeidx) = 1 \\
0 & \text{otherwise}
\end{cases}
\end{align}
Note that Conway's \emph{Game of Life} is turing-complete and its evolution is, in general,
unpredictable without computing every single step according to the rules \cite{Adamatzky2002}.
We created $30$ trajectories by initializing a $20 \times 20$ grid with one of six
standard patterns at a random position, namely \emph{blinker}, \emph{beacon}, \emph{toad},
\emph{block}, \emph{glider}, and \emph{block and glider} (see figure~\ref{fig:game_of_life}).
The first four patterns are simple oscillators with a period of two, the glider is an infinitely
moving structure with a period of two (up to rotation) and the \emph{block and glider} is a chaotic structure
which converges to a block of four and a glider after 105 steps. \footnote{Also refer to the \emph{Life Wiki}
\url{http://conwaylife.com/wiki/} for more information on the patterns.}
We let the system run for $T = 10$ time steps resulting in $300$ graphs overall. In every step,
we further activated $5\%$ of the cells at random, simulating observational noise.

\begin{figure}
\begin{center}
\includegraphics{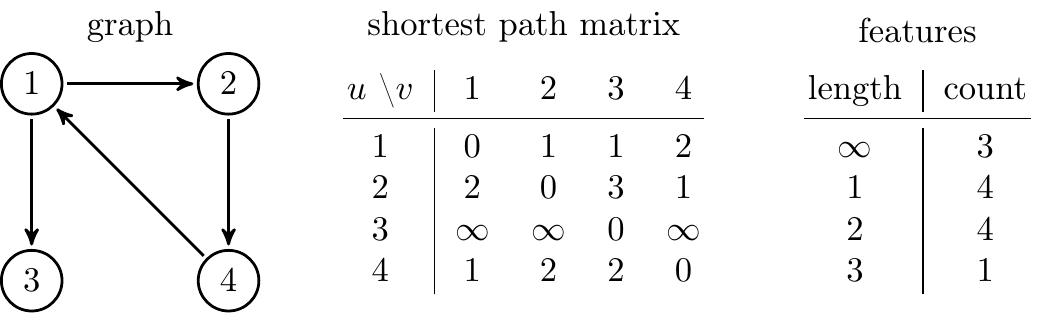}
\end{center}
\caption{An example graph, the associated matrix of shortest paths as returned by the Floyd-
Warshall algorithm \cite{Floyd1962} and the histogram over path lengths used as feature
representation for our approach. Note that self-distances are ignored.}
\label{fig:shortest_path_features}
\end{figure}

As data representation for the theoretical data sets we use an explicit feature embedding inspired
by the shortest-path-kernel of Borgwardt and colleagues \cite{Borgwardt2005}.
Using the Floyd-Warshall algorithm \cite{Floyd1962} we compute all shortest paths in the graph and
then compute a histogram over the lengths of these shortest paths as a feature map
(see figure~\ref{fig:shortest_path_features} for an example).
As dissimilarity we use the Euclidean distance on these feature vectors, which we normalize by the
average distance over the data set. We transformed the distance to a kernel via the radial basis function transformation.
Eigenvalue correction was not required.

\begin{table}
\caption{The mean RMSE and runtime across cross validation trials for both theoretical data sets
(x-axis) and all methods (y-axis). The standard deviation is shown in brackets. Runtime
entries with $0.000$ had a shorter runtime (and standard deviation) than $10^{-3}$ milliseconds.
The best (lowest) value in each column is highlighted by bold print.}
\begin{center}
\begin{tabular}{lcccc}
\multicolumn{1}{r}{} & \multicolumn{2}{c}{Barabási-Albert} & \multicolumn{2}{c}{Game of Life} \\
method   & RMSE                   & runtime [ms]           & RMSE                   & runtime [ms]\\
\cmidrule(lr){1-1} \cmidrule(lr){2-3} \cmidrule(lr){4-5}
identity & $0.137$ ($0.005$)      & $\bm{0.000}$ ($0.000$) & $1.199$ ($0.455$)      & $\bm{0.000}$ ($0.000$) \\
1-nn     & $0.073$ ($0.034$)      & $0.111$ ($0.017$)      & $1.191$ ($0.442$)      & $0.112$ ($0.025$) \\
KR       & $0.095$ ($0.039$)      & $0.122$ ($0.016$)      & $0.986$ ($0.398$)      & $0.120$ ($0.040$) \\
GPR      & $0.064$ ($0.028$)      & $0.148$ ($0.022$)      & $\bm{0.965}$ ($0.442$) & $0.127$ ($0.026$) \\
rBCM     & $\bm{0.062}$ ($0.015$) & $0.312$ ($0.083$)      & $0.967$ ($0.461$)      & $0.267$ ($0.077$) \\
\end{tabular}
\end{center}
\label{tab:results_theoretical}
\end{table}

The RMSE and runtimes for all three data sets are shown in table~\ref{tab:results_theoretical}.
As expected, KR, GPR and rBCM outperform the identity-baseline ($p < 10^{-3}$ for both data sets), supporting H1.
1-NN outperforms the baseline only in the Barabási-Albert data set ($p < 10^{-3}$).
Also, our results lend support to H2 as rBCM outperforms 1-NN in both data sets
($p < 0.05$ for Barabási-Albert, and $p < 0.01$ for Conway's \emph{Game of Life}).
However, rBCM is significantly better than KR only for the Barabási-Albert data set ($p < 0.001$),
indicating that for simple data sets such as our theoretical ones, KR might already provide a
sufficient predictive quality.
Finally, we do not observe a significant difference between rBCM and GPR, as expected in H3.
Interestingly, for these data sets, rBCM is slower compared to GP, which is probably due to
constant overhead for maintaining multiple models.

\subsection{Java Programs}

Our two real-world java data sets consist of programs for two different problems
from beginner's programming courses. The motivation for time series prediction on such data
is to help students achieve a correct solution in an intelligent tutoring system (ITS).
In such an ITS, students incrementally work on their program until they might get stuck and do not
know how to proceed. Then, we would like to predict the most likely next state of their
program, given the trajectories of other students who have already correctly solved the problem.
This scheme is inspired by prior work in intelligent tutoring systems, such as the hint factory
by Barnes and colleagues \cite{Barnes2008}.

As our data sets consist only of final, working versions of programs, we have to simulate
the incremental growth. To that end we first represented the programs as graphs via their
abstract syntax tree, and then recursively removed the last semantically important node 
(where we regarded nodes as important which introduce a visibility scope in the Java program,
such as class declarations, method declarations, and loops), until the program was empty. Then,
we reversed the order of the resulting sequence, as to achieve a growing program.
In detail, we have the following two data sets:

\begin{figure}
\includegraphics[width=\textwidth]{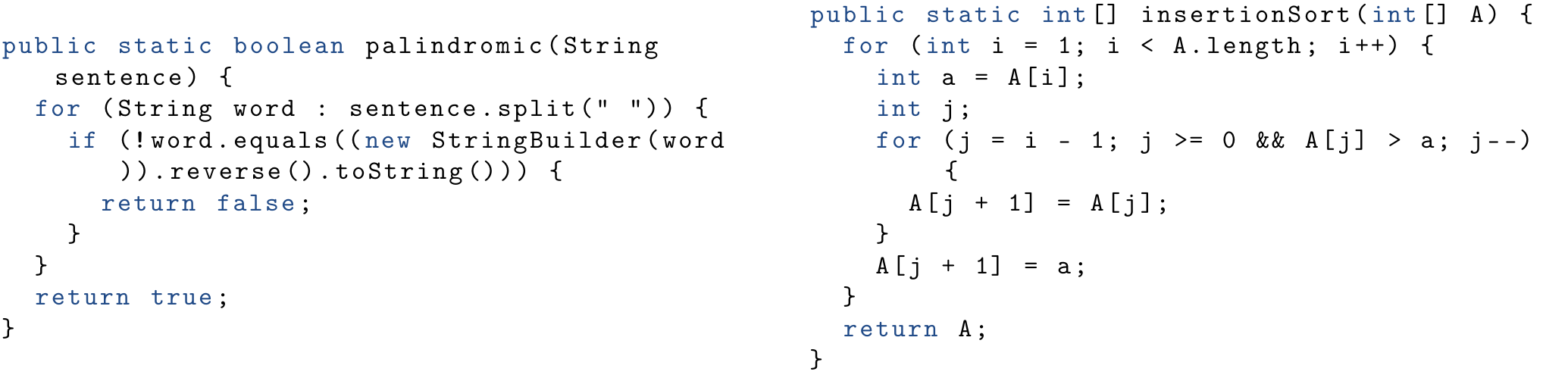}
\caption{Example Java code from the MiniPalindrome data set (left) and from the Sorting data set
(right).}
\label{fig:java}
\end{figure}

\paragraph{MiniPalindrome:} This data set consists of $48$ Java programs, each realizing one of
eight different strategies to recognize palindromic input (see figure~\ref{fig:java}) \cite{Mokbel2013} \footnote{The data
set is available online at \url{http://doi.org/10.4119/unibi/2900666}}. The abstract syntax trees
of these programs contain $135$ nodes on average. The programs come in eight different variations 
described in \cite{Mokbel2013}. Our simulation resulted in $834$ data points.

\paragraph{Sorting:} This is a benchmark data set of $64$ Java sorting programs taken from the web, 
implementing one of two sorting algorithms, namely \emph{BubbleSort} or \emph{InsertionSort}
(see figure~\ref{fig:java}) \cite{Mokbel2014}
\footnote{The data set is available online at \url{http://doi.org/10.4119/unibi/2900684}}.
The abstract syntax trees contain $94$ nodes on average. Our simulation resulted in $800$ data 
points.

To achieve a dissimilarity representation we first ordered the nodes of the abstract syntax
trees in order of their appearance in the original program code. Then, we computed
a sequence alignment distance on the resulting node sequences, similar to the method described by
Robles-Kelly and Hancock \cite{RoblesKelly2005}. In particular, we used an affine sequence alignment
with learned node dissimilarity as suggested in \cite{Paassen2016}. We transformed the dissimilarity
to a similarity via the radial basis function transformation and obtained a kernel via
clip Eigenvalue correction \cite{Gisbrecht2015}.

\begin{table}
\caption{The mean RMSE and runtime across cross validation trials for both Java data sets
(x-axis) and all methods (y-axis). The standard deviation is shown in brackets. Runtime
entries with $0.000$ had a shorter runtime (and standard deviation) than $10^{-3}$ seconds.
The best (lowest) value in each column is highlighted by bold print.}
\begin{center}
\begin{tabular}{lcccc}
\multicolumn{1}{r}{} & \multicolumn{2}{c}{MiniPalindrome} & \multicolumn{2}{c}{Sorting} \\
method   & RMSE                   & runtime [s]              & RMSE                   & runtime [s]\\
\cmidrule(lr){1-1} \cmidrule(lr){2-3} \cmidrule(lr){4-5}
identity & $0.295$ ($0.036$)      & $\bm{0.000}$ ($0.000$)   & $0.391$ ($0.029$)      & $\bm{0.000}$ ($0.000$)     \\
1-NN     & $0.076$ ($0.047$)      & $0.000$ ($0.000$)        & $0.090$ ($0.042$)      & $0.000$ ($0.000$)     \\
KR       & $0.115$ ($0.031$)      & $1.308$ ($0.171$)        & $0.112$ ($0.027$)      & $1.979$ ($0.231$)     \\
GPR      & $0.075$ ($0.064$)      & $111.417$ ($0.304$)      & $0.020$ ($0.034$)      & $114.394$ ($0.301$)   \\
rBCM     & $\bm{0.044}$ ($0.052$) & $11.698$ ($0.085$)       & $\bm{0.010}$ ($0.025$) & $18.5709$ ($0.121$)
\end{tabular}
\end{center}
\label{tab:results_java}
\end{table}

We show the RMSEs and runtimes for both data sets in table~\ref{tab:results_java}.
Contrary to the theoretical data sets before, we observe strong differences both between
the predictive models and the baseline, as well as between rBCM and 1-NN as well as KR ($p < 0.01$ in all cases),
which supports both H1 and H2. Interestingly, rBCM appears to achieve better results compared to GPR,
which might be the case due to additional smoothing provided by the averaging operation over all
cluster-wise GPR results. This result supports H3.
Finally, we observe that rBCM is about 10 times faster compared to GPR.

\section{Discussion and Conclusion}

Our results indicate that it is possible to achieve time series prediction in kernel
and dissimilarity spaces, in particular for graphs. In all our experiments, even simple
predictive models (1-nearest neighbor and kernel regression) did outperform the baseline of staying where you are.
For real-world data we could further improve the predictive error significantly by applying a more complex predictive
model, namely the robust Bayesian Committee Machine (rBCM).
This indicates a trade-off in terms of model choice: Simpler models are faster
and in the case of 1-nearest neighbor the result is easier to interpret, because it is given as a graph.
However, in the case of real-world data, it is likely that a more complex predictive model is required
to accurately describe the underlying dynamics in the kernel space. Fortunately, the runtime
overhead is only a constant factor as rBCM can be applied in linear time.

Our key idea is to apply time series prediction in a (pseudo-)Euclidean space and representing
the graph output as a point in this space. Even though we do not know the graph corresponding to the
predicted location in the latent space, we have shown that this point can be analyzed
using subsequent dissimilarity- or kernel-based methods as the dissimilarities and kernel values
with respect to the predicted point can still be calculated.

However, a few challenges for further research remain. First, usual hyperparameter optimization techniques
for Gaussian processes depend on a vectorial data representation \cite{Deisenroth2015} and thus are not
necessarily applicable in our proposed pipeline. Therefore, alternative hyperparameter selection techniques
are required.
Second, theoretic or empirical results regarding the number of data required to make accurate predictions
are still lacking for this novel approach.
Finally, for some applications, the predicted point in the primal space may be required, that is,
we need a prediction in form of a graph, for example for feedback provision in intelligent tutoring
systems or predicting the precise structure of a sensor network, e.g.\ for predictive maintenance
in an oil refinery.
For such cases,  we have to solve a pre-image problem: Finding the original point that maps to the affine 
combination in the pseudo-Euclidean space. This problem has proved particularly challenging in the
literature up until now and could profit from further consideration \cite{bakir_learning_2003,bakir_learning_2004,Kwok2004}.

\bibliographystyle{abbrvurl}
\bibliography{literature.bib}

\end{document}